%% file: main.tex
\documentclass[]{alaya}
\usepackage{makecell}
\usepackage{wrapfig}
\usepackage{tabularx}
\usepackage{textcomp}
\usepackage{stfloats}
\usepackage{url}
\usepackage{verbatim}
\usepackage{titlesec}
\usepackage{adjustbox}
\usepackage{multirow}
\usepackage{pifont}
\usepackage[sc]{mathpazo}
\usepackage{tikz}
\usepackage{amsmath,amssymb}
\usepackage{colortbl}
\usepackage[numbers,sort&compress]{natbib}
\usepackage{booktabs}
\usepackage{hyperref}
\usepackage{graphicx}
\RequirePackage{xspace}
\makeatletter
\DeclareRobustCommand\onedot{\futurelet\@let@token\@onedot}
\def\@onedot{\ifx\@let@token.\else.\null\fi\xspace}
\usepackage[most]{tcolorbox}
\usepackage{array}
\usepackage{siunitx}
\usepackage{caption}
\definecolor{headerpurple}{HTML}{d8d2fc}
\definecolor{rowgray}{gray}{0.95}
\usepackage{CJKutf8}

\makeatother

\definecolor{adptorange}{RGB}{248, 205, 172}
\definecolor{cmpblue}{RGB}{189, 215, 238}

\definecolor{our_red}{RGB}{232,157,160}
\definecolor{our_blue}{RGB}{136,206,230}
\definecolor{our_orange}{RGB}{246,200,168}
\definecolor{our_green}{RGB}{178,211,164}

\definecolor{mygray}{HTML}{f0f0f0}

\usepackage{bbding}
\usepackage{fontawesome}
\usepackage{float}
\usepackage{flafter}
\usepackage{needspace}
\usepackage{algorithm}
\usepackage{algorithmic}
\usepackage{listings}
\definecolor{lstbg}{gray}{0.96}
\definecolor{lstcomment}{rgb}{0.35,0.45,0.40}
\newlength\savewidth

\newcolumntype{x}[1]{>{\centering\arraybackslash}p{#1pt}}
\newcolumntype{y}[1]{>{\raggedright\arraybackslash}p{#1pt}}
\newcolumntype{z}[1]{>{\raggedleft\arraybackslash}p{#1pt}}

\renewcommand{\paragraph}[1]{\vspace{1.25mm}\noindent\textbf{#1}}

\definecolor{green}{HTML}{009000}
\definecolor{red}{HTML}{ea4335}


\title{DirCast: Live Esports Directing and Commentary with a Single Streaming Video-Language Model}

\author[1,2]{\mbox{Zixian Gao}}
\author[2]{\mbox{Lixin Xiu}}
\author[3]{\mbox{Penghao Luo}}
\author[1]{\mbox{Xiangcheng Cao}}
\author[1,4,5]{\newline\mbox{Ziqi Cai}}
\author[1,2]{\mbox{Zhen Li}}
\author[5]{\mbox{Erwin Wu}}
\author[1]{\mbox{Kaipeng Zhang}}
\affiliation[1]{Alaya Lab}
\affiliation[2]{The University of Tokyo}
\affiliation[3]{\mbox{Georgia Institute of Technology}\newline}
\affiliation[4]{Peking University}
\affiliation[5]{\mbox{Institute of Science Tokyo}}

\input{sections/abstract}

\github{\url{https://zixiangao.github.io/DirCast/}}
\Code{\url{https://github.com/ZixianGao/DirCast-code}}
\correspondence{Kaipeng Zhang}
\date{\today}

\begin{document}
\maketitle

\vspace{10pt}
\begingroup
    \centering
    \includegraphics[width=\linewidth]{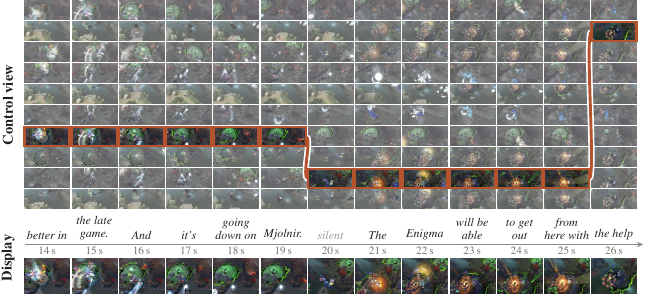}\vspace{5pt}
    \captionsetup{hypcap=false}
    \captionof{figure}{DirCast directs and commentates a live Dota 2 match. Top: the ten hero cameras over 13 seconds, with the on-air camera framed. Bottom: the program that goes on air and the words DirCast speaks over it.}
    \label{fig:teaser}
\endgroup

\input{sections/introduction}

\input{sections/related_work}
\input{sections/dataset}
\input{sections/method}
\input{sections/evaluation}

\input{sections/conclusion}

\bibliographystyle{plainnat}
\bibliography{references}

\beginappendix
\input{sections/appendix_dataset}
\input{sections/appendix_qualitative}

\end{document}

%% file: sections/abstract.tex
\abstract{
A live esports broadcast needs a director who chooses which camera goes on air and a caster who talks over that choice, every second and without knowing what comes next. Streaming video language models have learned to commentate, but only over a broadcast that a human director has already produced. We study live broadcast production, in which a model receives the raw camera of every player and must both direct and commentate from the past alone. We present DirCast, a single streaming video language model that does both. Every second it scans all cameras at low resolution, decides with a hold-or-switch director whether and where to cut, and reads the chosen camera at high resolution before it speaks. Its context keeps what went on air and what was said and lets the other views expire, which bounds the cost of a stream of any length. We build a dataset of 27 professional Dota 2 matches with a camera for every hero, the human director's choices and the casters' commentary. DirCast directs better than rule-based and zero-shot directors including GPT-5, and its commentary wins 29.2 percent of comparisons against GPT-4o under the Proact-VL protocol, against 19.4 percent for the strongest of eleven streaming commentators that watch the human broadcast.
}

%% file: sections/introduction.tex
\section{Introduction}
\label{sec:intro}

A live esports broadcast is produced by two people working in tight coordination. The director watches a wall of camera feeds, one per player, and decides every second which feed goes on air. The caster watches what the director puts on air and talks over it, naming the heroes, calling the fights, and filling the quiet moments with analysis. Neither job can wait for the future. When a fight breaks out on the far side of the map, the director has a few seconds to cut to it, and the caster has to describe it while it is still happening. This is why professional broadcasts need trained crews, and why the long tail of matches, from regional leagues to amateur games, is streamed with a fixed camera and no commentary.

Recent video language models~\cite{li2024llava,wang2025internvl3,bai2025qwen3} have started to take over the second job. Streaming commentators such as LiveCC~\cite{chen2025live}, StreamingVLM~\cite{xu2026streamingvlm} and Proact-VL~\cite{yan2026proact} read a video one frame at a time and decide when to speak and what to say~\cite{chen2024videollm,qian2025dispider,wang2025videollm,wang2026mmduet2,yang2026livestar}. They all assume, however, that the first job has already been done. Their input is a finished broadcast in which a human director has chosen every shot. The model inherits the director's judgment for free, and the hardest part of live production, deciding what the audience should see, is left outside the problem. A system that is given only the raw feeds of a match cannot use these models at all.

We study the full problem. Given the ten hero cameras of a live Dota 2 match and the game events as they happen, a model has to choose the on-air feed every second and produce live commentary over its own choice, using only the past. We call this task live broadcast production. It couples two decisions that are usually studied apart~\cite{chen2018camera,chen2019learning,zhang2022moba,rao2024matchtime}. The director must track ten simultaneous scenes and recognize which one is about to matter~\cite{chen2016learning,majumder2025viewpoint}. The commentator must describe a picture that changes whenever the director cuts, and its words are only good if the director put the right action on screen. A model that solves the task produces a complete broadcast from raw input.

The coupling makes the task hard in ways that neither half shows alone. Ten feeds at one frame per second produce more visual tokens than a streaming model can keep in context for long~\cite{song2024moviechat,zhang2024flash,qian2024streaming,di2025streaming}. The decisions compound over time, because a director that misses a fight shows the commentator the wrong scene, and every later second is conditioned on that early mistake. Training on human broadcasts does not prepare the model for this, since at training time it only ever sees the program a human chose, the classic exposure bias~\cite{bengio2015scheduled,ross2011reduction}. Human directors also hold a shot most of the time and cut rarely, so a model trained to imitate them learns to never cut at all. And a streaming model that is allowed to stay silent tends to fall into silence and never fully recover.

We present DirCast, a single causal video language model that directs and commentates at once. Every second, the model reads a short chunk that contains the new game events and the ten feeds at low resolution, each paired with a token that encodes the hero's game state. A hold-or-switch director head then decides whether to keep the current shot or to cut, and a pointer over the ten feeds chooses where to cut to. The chosen feed is appended to the stream at high resolution, and the commentary is generated after it, so the commentator always speaks over the shot the director chose. Low-resolution feed tokens leave the context after a few seconds, while the on-air views and the spoken words stay. The memory of the model is the program it has produced, which keeps the cost per second bounded and lets DirCast run in real time on a single GPU.

Three further choices make the model robust when it runs on its own. We roll out perturbed programs during training, in which the on-air feed drifts away from the human choice for a few seconds while the targets stay human, so the director learns to recover from its own mistakes and the commentator learns to speak over any shot. We ground every kill in the event stream to the feed of the hero who made it, and we accept a cut to a fight participant as a correct directing decision, which teaches the director that action is worth showing even when the human chose otherwise. At inference, the thresholds for cutting and for speaking follow the running rates of the model's own decisions toward the rates of human crews, which keeps the program lively and the commentary present without any tuning on test data.

To study the task we build a dataset from 27 professional Dota 2 matches. We render one camera per hero from the match replays, align the renders with the official broadcast, and label every second with the feed the human director put on air and the words the casters spoke. We evaluate commentary with the protocol of Proact-VL~\cite{yan2026proact} and direct comparison against eleven streaming commentators, and we evaluate directing by agreement with the human director and by coverage of the kills in the match. DirCast wins 29.21 percent of its comparisons against GPT-4o, while the strongest streaming commentator wins 19.38 percent, even though every baseline is given the finished human broadcast and DirCast directs its own. It starts speaking 0.30 seconds after the human casters on average. As a director, it agrees with the human crew more often than every rule-based and zero-shot director, it covers 89.1 percent of kills, and it cuts at a rate close to the human crew. The best zero-shot vision language model agrees with the human crew only two points more often than a static director that never cuts, and GPT-5 agrees less often. Trained as one model, DirCast is also better at both roles than two models trained separately, with half the memory.

Our contributions are as follows.
\begin{itemize}
\item We introduce live broadcast production, directing and commentating a live match from raw camera feeds, with a dataset of 27 professional Dota 2 matches.
\item We propose DirCast, a single streaming video language model that decides what goes on air with a hold-or-switch director and comments on the shot it chooses, in real time.
\item DirCast outperforms eleven streaming commentators while directing its own broadcast, and directs better than rule-based and zero-shot directors, including GPT-5.
\end{itemize}

%% file: sections/related_work.tex
\section{Related Work}
\label{sec:related}

\paragraph{Streaming video language models.}
Video language models such as Video-LLaVA~\cite{lin2024video}, LLaVA-OneVision~\cite{li2024llava}, InternVL3.5~\cite{wang2025internvl3}, Qwen2.5-VL and Qwen3-VL~\cite{bai2025qwen3} process prerecorded clips, whereas streaming models must respond as video arrives. VideoLLM-online~\cite{chen2024videollm} learns when to remain silent, while subsequent work uses separate decision modules~\cite{qian2025dispider,wang2026streambridge}, interleaved dialogue~\cite{wang2025videollm,wang2026mmduet2} or learned response timing~\cite{yang2026livestar}. Long-term context is managed through memory banks~\cite{song2024moviechat,he2024ma}, compression~\cite{zhang2024flash,qian2024streaming,di2025streaming}, KV-cache retrieval~\cite{di2025streaming}, recent-window attention~\cite{xiao2024efficient}, token eviction~\cite{zhang2023h2o} and compact frame representations~\cite{li2024llama,xu2024slowfast}. StreamingBench~\cite{lin2026streamingbench} and OVO-Bench~\cite{niu2025ovo} evaluate online understanding. For live commentary, LiveCC~\cite{chen2025live} learns from large-scale speech supervision, StreamingVLM~\cite{xu2026streamingvlm} supports hours of commentary with bounded memory, and Proact-VL~\cite{yan2026proact} introduces proactive gaming commentary and an evaluation protocol we adopt. These methods assume a preselected video stream. DirCast instead produces that stream through causal directing and maintains role-aware memory of aired views and spoken commentary.

\paragraph{Commentary for sports and games.}
Automated commentary spans chess~\cite{jhamtani2018learning}, online video~\cite{ma2019livebot} and soccer, where SoccerNet~\cite{giancola2018soccernet} supports event captioning~\cite{mkhallati2023soccernet}, knowledge-grounded commentary~\cite{qi2023goal} and timeline-aligned narration~\cite{rao2024matchtime}. Games provide structured records for commentary generation, including racing with video and telemetry~\cite{ishigaki2021generating} and esports datasets pairing League of Legends or Dota 2 gameplay with professional commentary~\cite{tanaka2021lol,zhang2022moba,zhang2024game}. These approaches rely on a single produced broadcast or game data without video. Our dataset preserves all available camera views, enabling joint training and evaluation of shot selection and commentary.

\paragraph{Automatic directing and view selection.}
Automatic cinematography includes editing multi-camera social footage~\cite{arev2014automatic}, learning dialogue editing conventions~\cite{leake2017computational} and assembling videos from text~\cite{wang2019write}. Sports systems learn camera control~\cite{chen2016learning} and view selection based on visual importance~\cite{chen2018camera,chen2019learning}. Related work selects viewpoints in panoramic video~\cite{su2016pano2vid,hu2017deep} or informative views in multi-camera instructional recordings~\cite{majumder2025viewpoint,grauman2024ego}. Unlike offline view selection, DirCast directs causally at each second and couples each decision with commentary. Its coarse-to-fine processing resembles glimpse-based attention~\cite{mnih2014recurrent}. Since directing decisions alter subsequent observations, the model also faces exposure bias, studied through scheduled sampling~\cite{bengio2015scheduled}, professor forcing~\cite{lamb2016professor} and expert-corrected imitation learning~\cite{ross2011reduction}. Our perturbed program rollouts similarly expose the director to states induced by its own errors.

%% file: sections/dataset.tex
\section{Dataset}
\label{sec:dataset}

Live broadcast production needs data that no existing video corpus provides. A broadcast shows one camera at a time, and the cameras the director did not choose are lost. To learn directing, a model has to see every camera the director could have chosen, together with the one the director actually put on air and the words the casters spoke over it.

Esports make this possible. A professional Dota 2 match is fully described by its replay file, so the game can be played back and filmed from any point of view after the fact. We collect 27 recent professional matches with their official broadcasts. For every match we render ten cameras from the replay with the game's own spectator client, one following each hero. We then align the rendered timeline with the broadcast and label every second with three complementary signals. The first is the on-air camera, the hero feed that best matches what the broadcast shows at that second. The second is the state of the broadcast, which separates live gameplay from replays, overlays and breaks, so that only live seconds count as directing decisions. The third is the commentary, transcribed from the broadcast audio with word timestamps and cleaned following the Proact-VL polishing pipeline, which corrects misheard hero and player names while keeping the timing of every word. From the replay we further extract the state of every hero at every second, such as health, nearby enemies and recent damage, and the stream of game events, such as kills and destroyed buildings, each placed at the second it happens.

We cut the aligned matches into clips of 36 seconds at one frame per second, the setting used by streaming commentary benchmarks. The split is by match, with 20 matches for training, 3 for calibration and 4 for testing, so that no team, player or game of the test set is seen in training. The training set contains 2,096 clips and 21 hours of play, and the test set contains 496 clips and 5 hours. Human directors cut 7.9 times per minute on average and hold most shots for several seconds. The casters speak in about three quarters of all seconds. These two numbers describe the rhythm a model has to reproduce, and we return to them when we discuss inference.

%% file: sections/method.tex
\begin{figure}[t]
    \centering
    \includegraphics[width=\linewidth]{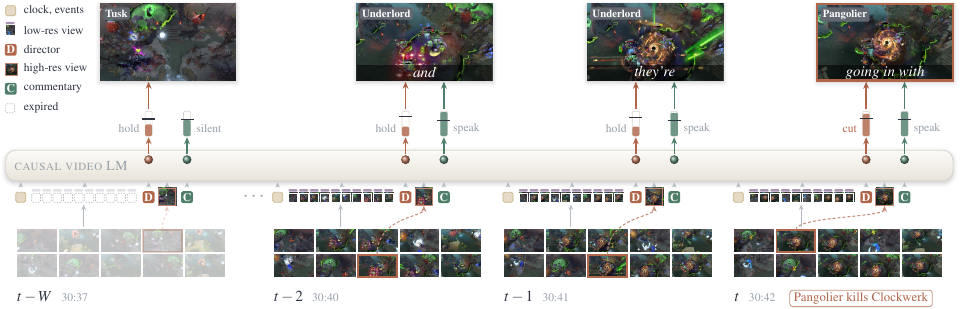}
    \caption{Framework. Each second a low-res ViT encodes the ten feeds with their state, the director holds or cuts, the chosen feed is read at high resolution, and the commentator speaks. Low-res views expire while the program stays.}
    \label{fig:method}
\end{figure}
\section{Method}
\label{sec:method}

Every second, DirCast receives the ten camera frames and the new game events, decides which camera goes on air, and decides whether and what to say. It may only use the past. We build the model around one observation. A director and a commentator are not two separate systems that happen to run at the same time. The director decides what the audience sees, the commentator talks about what the audience sees, and both depend on what has been shown and said so far. We therefore use a single causal video language model for both roles, in which the commentary is generated after, and conditioned on, the directing decision. Figure~\ref{fig:method} gives an overview.

\subsection{One stream for directing and commentary}

We start from a pretrained video language model and feed it the match as a single token stream that grows by one chunk per second. The chunk for second $t$ has a fixed layout. It opens with the game clock and the events of that second. It then lists the ten cameras, each as a low-resolution view $v^k_t$ preceded by a state token $s^k_t$ that embeds the game state of hero $k$. A director token $\langle\mathrm{D}\rangle$ follows. After the directing decision, the chosen camera $c_t$ is appended again at high resolution, followed by a commentary token $\langle\mathrm{C}\rangle$ and the words $w_t$ spoken in that second.

The order of the chunk follows the order in which a human crew works. The director scans all cameras at a glance, picks one, and the commentator then looks closely at the picture that went on air. Each step reads everything before it, so the director sees the full history of the broadcast and the commentator sees the director's choice. The two resolutions keep the stream affordable. Ten cameras at full resolution would produce far more visual tokens than a streaming model can process every second, and most of them show nothing worth describing. A low-resolution view is enough to tell where the action is, while the commentator needs a sharper view to name heroes, items and abilities. DirCast therefore reads only the camera that goes on air at high resolution.

\subsection{Hold or switch}

Most of the time a good director does nothing. Shots are held for several seconds, and a cut is a deliberate event that happens when something more important appears elsewhere. A classifier over ten cameras does not capture this structure. It must relearn at every second that the current camera is usually the right one, and small changes in its scores produce flickering cuts.

We instead factor the decision into whether to cut and where to cut to. Let $h_{\mathrm{D}}$ be the hidden state of the director token and let $\bar h_k$ be the mean hidden state of the tokens of camera $k$. A cut gate
\begin{equation}
g_t = \sigma\big(f([h_{\mathrm{D}}\,,\,\bar h_{c_{t-1}}]) + \psi(s_t)\big)
\end{equation}
estimates the probability of leaving the camera that is on air, given the director state and the current shot. A pointer
\begin{equation}
\pi_t(k) \propto \exp \big\langle q(h_{\mathrm{D}})\,,\,\kappa(\bar h_k) + e_k + \phi(s^k_t) \big\rangle , \quad k \neq c_{t-1},
\end{equation}
scores the other cameras by comparing the director state with each camera's content and game state, where $e_k$ is a learned embedding of the player slot. The probability of each camera is then
\begin{equation}
P(c_t = k) = (1-g_t)\,\mathbf{1}[k = c_{t-1}] + g_t\,\pi_t(k).
\end{equation}
Holding is the default, and switching requires the gate to open. The pointer treats all cameras with the same function, so it ranks cameras by what they show instead of by their position in the input.

\subsection{The program is the memory}

A streaming model must forget, or its context grows without bound. The question is what to keep. A human crew does not remember what every camera showed a minute ago. It remembers the broadcast, the shots that actually went on air and what was said over them. DirCast keeps this. The low-resolution views of the ten cameras stay for only $W$ seconds and then leave the context, while the high-resolution on-air views, the events and the words stay. During training this rule is an attention mask, and during streaming the expired entries are removed from the key-value cache. The model's memory is the program it has produced. On streams longer than a training clip, every second also leaves the cache once it is older than a clip, so the cost per second stays constant over a match of any length.

\subsection{Grounding the game}

Video alone is a weak signal for directing. A fight that is about to start looks like two groups of heroes walking, and a kill on the far side of the map is a small change in one of ten thumbnails. The game itself knows more. The state token of each camera gives the model the hero's health, recent damage and the enemies around it, and the event text states what happened, such as one hero killing another. We tie each event to the camera that shows it, so that the kill of a hero raises a flag on the camera of the hero who made it. The model thus reads the game in the same terms as the cameras it has to choose from.

\subsection{Learning to direct from human broadcasts}

DirCast is trained on human broadcasts with three objectives. A language modeling loss teaches the commentary, a binary loss on the commentary token teaches when to speak, and a directing loss teaches the gate and the pointer. The gate learns from the human cuts, and the pointer learns which camera the human chose at every cut, where a camera whose hero is visible in the broadcast shot counts as a correct answer as well.

Imitating human broadcasts directly leaves two gaps. The first gap is between training and deployment. During training the model always sees the program the human chose. At test time it sees its own program, and a single wrong cut moves it to a state it has never seen. We close this gap with on-policy program rollouts. During training, the on-air camera in the input drifts away from the human choice for a few seconds at a time, while the targets stay human. The director learns to recover from a wrong shot by cutting back, and the commentator learns to describe whatever is on air. The second gap is in the targets. A human director makes one choice among several good ones. When a kill happens on a camera that is not on air, cutting to it is a correct directing decision even if the human stayed elsewhere. We add an event-anchored target that, in the seconds around a kill, also accepts the cameras of the killer and the victim, and asks the gate to open when neither is on air. The model learns that action is worth showing, which is the part of directing that the human labels express least consistently.

\subsection{Streaming decisions at a human rhythm}

At inference, the model cuts when the gate exceeds a threshold and speaks when the probability of the commentary token exceeds a threshold. Fixed thresholds fail in streaming. A model that stays silent for a few seconds conditions on its own silence and becomes more likely to stay silent, and a director that has not cut for a while drifts the same way. We let both thresholds follow the model’s own behavior,
\begin{equation}
\tau_t = \tau_0 + \gamma\,(\bar r_t - r^{*}),
\end{equation}
where $\bar r_t$ is a running average of the model's recent decisions and $r^{*}$ is the rate of human crews measured on the training set. When the model has been quieter than human casters, the threshold to speak drops, and when it has been cutting more often than human directors, the threshold to cut rises. The running rate of speech starts at zero, since the model has not spoken when it joins a stream. Once the model starts an utterance, it ends it only when its end probability is high, so that each second carries a complete phrase rather than a fragment. These rules carry the rhythm of a human broadcast into the model without any training signal and keep the broadcast lively for as long as it runs.

%% file: sections/evaluation.tex
\section{Experiments}
\label{sec:experiments}

\subsection{Setup}
\label{sec:setup}

\paragraph{Protocol.}
All results are on the test split of four unseen matches, 496 clips with 485 commented and 11,283 strict LIVE seconds. Every model runs free through every second of each clip and conditions only on its own past decisions.

\paragraph{Directing.}
We measure how often a model agrees with the human director (Top-1, Tol-top1 with one second of slack, and Any-visible, which accepts any feed whose hero is visible on air). We also report the precision and F1 of correct cuts, and the coverage of kills and fights on air. We compare with three rules and eight zero-shot VLMs. The random rule cuts to a random feed at the human cutting rate, and the static rule keeps one feed throughout. The action rule scores every feed by the game state of its hero, nearby enemies, recent damage, deaths and crowding, and cuts to the highest-scoring feed when it leads and the current shot has been held at least three seconds. The zero-shot models~\cite{bai2025qwen3,wang2025internvl3,li2024llava} include proprietary GPT-5 and Kimi-K2.6 and see the same feeds, heroes and events as DirCast.

\input{sections/tables/direct}

\paragraph{Commentary.}
We follow the Proact-VL evaluation protocol~\cite{yan2026proact} and run its publicly released scripts. CC is the win rate against offline GPT-4o commentary, F1 and TimeDiff measure whether and when the model chooses to speak, LiveU and FinalQ rate the live delivery and the quality of the final transcript, and PAUC summarizes commentary quality over the whole stream. GPT-5.1, the official judge, is not available to us, so all models, including ours, are judged by GPT-5 instead. We compare with eleven streaming commentators~\cite{chen2025live,xu2026streamingvlm,qian2025dispider,wang2025videollm,wang2026mmduet2,yang2026livestar,yan2026proact}. Nine run zero-shot with their released weights, and LiveCC-7B and Proact-VL are also fine-tuned on our training split. Each of them watches the finished human broadcast rather than choosing its own shots from the feeds.

\paragraph{Implementation.}
DirCast fine-tunes Qwen3-VL-8B~\cite{bai2025qwen3} with LoRA for three epochs on eight H100 GPUs, using the language modeling, speaking and directing losses of Section~\ref{sec:method} together with perturbed program rollouts. The model reads the ten feeds at 128 pixels and the on-air feed at 288 pixels, and low-resolution views expire after $W=8$ seconds. Inference runs on a single GPU.

\subsection{Main Results}
\label{sec:results}

\paragraph{Directing.}
Table~\ref{tab:director} compares DirCast with rule-based and zero-shot directors. DirCast is the best director on every measure of agreement with the human crew. It reaches a Top-1 of 26.95 and an Any-visible of 41.81, against 25.43 and 36.11 for the action rule, the strongest baseline, and it has the highest precision and F1 of correct cuts. It covers 89.1 percent of kills, more than any baseline and more than the human director, who often stays with a shot when a kill happens elsewhere. Zero-shot vision language models do not direct well, whatever their size. None of the eight reaches the action rule on agreement, scale does not help within a family, and GPT-5 reaches a Top-1 of only 13.53. They also lack a sense of rhythm. Most of them cut 20 to 56 times per minute, against 7.9 for the human crew, so that most of their cuts are wrong and their cut precision stays between 6 and 9 percent. DirCast cuts 6.7 times per minute. The action rule follows fights by construction and has the highest fight coverage at 62.4. DirCast reaches 41.1, ahead of every zero-shot model and of the human director at 37.4, even though it is trained to imitate that director.

\input{sections/tables/commentary}

\paragraph{Commentary.}
Table~\ref{tab:commentary} compares DirCast with nine zero-shot streaming commentators and LiveCC fine-tuned on our data. DirCast wins 29.21 percent of its comparisons against GPT-4o, ten points above the strongest baseline, LiveCC-7B-Base at 19.38, and each seed beats LiveCC-7B-Base and Proact-VL with $p<0.001$ under a paired bootstrap over clips. The lead holds although every baseline watches the human broadcast while DirCast comments on its own program. DirCast also starts speaking closest to the human casters, with a TimeDiff of 0.30 seconds, and leads all three live delivery scores, the continuity and substance of the final transcript, and PAUC, with 0.024 against 0.015 for the best baseline. The two remaining metrics instead reward an extreme speaking rate. The casters speak in 73.6 percent of the seconds, so speaking in every second already reaches an F1 of 84.8. Proact-VL speaks in 99.9 percent and obtains the highest F1 of 83.21, while DirCast speaks in 86.6 percent and obtains 78.22. Fidelity favors the opposite extreme. MMDuet speaks in 10.1 percent and so makes few claims that could be wrong. Among models that speak in more than half of the seconds, DirCast has the highest fidelity at 4.97, ahead of Proact-VL at 4.54.

\subsection{Ablation Studies}
\label{sec:ablation}
\input{sections/tables/abl_arc}
\paragraph{One model or two.}
Table~\ref{tab:joint} pairs DirCast with copies trained for a single role, each commentator speaking over its director's program. The joint model is better at both roles. It directs with a Top-1 of 27.00 against 25.17, wins 29.21 percent of comparisons against 24.36, and halves the memory to 18.9 GB by sharing weights. To check that the gain does not come from a better program, we compare both commentators on the human program, where the joint model still wins 27.84 percent against 25.67. Learning to direct thus makes the model a better commentator.

\input{sections/tables/componets}
% \paragraph{What makes a director.}
% Table~\ref{tab:ablation} ablates the directing components. Removing the hold-or-switch pointer costs the most, dropping Top-1 from 27.00 to 21.65 and Any-vis.\ from 41.80 to 35.93, which confirms that an explicit hold-or-switch decision is the backbone of the director. Removing state tokens or program rollouts also hurts, dropping Top-1 to 22.59 and 22.90, showing that game state complements vision and that training under self-induced states matters. Program rollouts help the commentary as well: without them CC falls from 29.21 to 28.35, since a commentator that never sees its own drifted program is less prepared for the shots the director actually produces. Without event-anchored targets, Top-1 is comparable but Any-vis.\ rises from 41.80 to 43.06, a trade-off between matching the human director and covering every visible event on air. Commentary metrics are otherwise stable throughout, so these components mainly affect directing.
\paragraph{What makes a director.}
Table~\ref{tab:ablation} ablates the directing components. Removing the hold-or-switch pointer costs the most, dropping Top-1 from 27.00 to 21.65 (a relative loss of about 20\%) and Any-vis.\ from 41.80 to 35.93, which confirms that an explicit hold-or-switch decision is the backbone of the director. Without it, the director must re-select a camera from scratch at every step, which likely leads to unmotivated switches. Removing state tokens or program rollouts also hurts, dropping Top-1 to 22.59 and 22.90 and Any-vis.\ to 37.98 and 38.01, showing that game state complements vision and that training under self-induced states matters. Program rollouts help the commentary as well: without them CC falls from 29.21 to 28.35, since a commentator that never sees its own drifted program is less prepared for the shots the director actually produces. Removing state tokens, by contrast, leaves CC unchanged at 29.21, so their benefit is confined to directing. Without event-anchored targets, Top-1 is comparable but Any-vis.\ rises from 41.80 to 43.06, a trade-off between matching the human director and covering every visible event on air. Since Top-1 is our primary measure of agreement with human directors, we keep event-anchored targets in the full model. Commentary metrics are otherwise stable throughout, with TimeDiff staying within 0.30--0.32, so these components mainly affect directing.

\subsection{Further Analysis}
\label{sec:analysis}

\input{sections/tables/memory}
\paragraph{Long streams.}
% Table~\ref{tab:longstream} streams eight continuous windows of 288 seconds each, eight times the length of a training clip. Keeping every feed view is the most expensive option: the cache reaches 92.6k tokens and 32.0\,GB, and the P95 cost rises to 1.05 seconds per video second, so the slowest seconds fall behind real time. Program memory, which retains only the views put on air, roughly halves this growth, cutting the final cache to 47.6k tokens and the P95 cost to 0.68 seconds. A horizon reduces cost further: 72 seconds already brings the cache down to 12.8k tokens, and 36 seconds, the length of a training clip, keeps it at 6.9k tokens and 18.9\,GB. The cost per video second then stays flat at 0.29 seconds, identical over the whole stream and its last 36 seconds, so the model can run indefinitely on a single GPU without slowing down. The shorter context does not hurt directing: Top-1 rises from 23.7 to 25.1, suggesting that stale views from minutes earlier mostly distract the director rather than inform it. The horizon also keeps the commentary active late in the stream, where the speaking rate otherwise decays, raising Speech from 67.2 to 76.7. Since every test clip is exactly 36 seconds long, the horizon never removes any tokens during standard evaluation, and all results in Tables~\ref{tab:director} and~\ref{tab:commentary} are unchanged.
Table~\ref{tab:longstream} streams eight continuous windows of 288 seconds, eight times the length of a training clip. Keeping every feed view doubles the cache to 92.6k tokens and pushes the slowest seconds beyond real time, with a P95 cost of 1.05 seconds per video second. Program memory halves the growth to 47.6k tokens. A 72-second horizon cuts the cache further to 12.8k tokens and the P95 cost to 0.58 seconds. A horizon that drops every token older than 36 seconds, the length of a training clip, keeps the cache at 6.9k tokens, memory at 18.9 GB, and the cost per second constant at 0.29 seconds on one GPU. The shorter context does not hurt direction, as Top-1 accuracy rises from 23.7 to 25.1. The horizon also keeps the commentary alive late in the stream, where the speaking rate otherwise decays, lifting speech from 67.2 to 76.7. Since every test clip is 36 seconds long, the horizon leaves all results in Tables~\ref{tab:director} and~\ref{tab:commentary} unchanged.

\begin{wrapfigure}[14]{r}{0.40\textwidth}
    \centering
    \includegraphics[width=\linewidth]{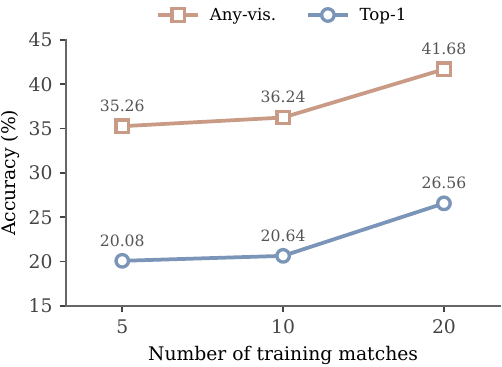}
    \caption{Accuracy vs.\ training matches.}
    \label{fig:scale}
\end{wrapfigure}
\paragraph{Scaling with training data.}
We train DirCast on 5, 10 and 20 matches and evaluate the director on the same test split (Figure~\ref{fig:scale}). Top-1 counts a second as correct when the model airs the same feed as the human director, and Any-vis.\ when the chosen feed shows the event at all. Going from 5 to 10 matches gives small gains of 0.98 in Any-vis.\ and 0.56 in Top-1, while doubling to 20 gives 5.56 and 6.36, reaching 41.80 and 27.00. The jump suggests that the director needs to see enough fights, teams and strategies before its choices generalize to unseen matches. The curves show no sign of saturation, and new matches are cheap to collect since each needs only its camera feeds and the aired broadcast, with no manual labels, so scaling to hundreds of matches is a natural next step. Because the subsets are nested and the test split is fixed, the rising curve reflects added training variety rather than an easier evaluation.

% \input{sections/tables/one_two}
% \paragraph{One model or two.}
% Table~\ref{tab:joint} compares DirCast with a pipeline of two models of the same size, a director and a commentator trained separately, in which the commentator comments on the program of the director. The pipeline directs better by 1.8 points of Top-1, and its commentary reaches a CC of XX against 29.28. It needs twice the parameters, XX GB of GPU memory against 18.9 and XX seconds of compute per second of video against 0.29. A single model thus produces the whole broadcast at about half the cost of two, with a small loss in directing.

\begin{figure}[htb]
    \centering
    \includegraphics[width=1.0\linewidth]{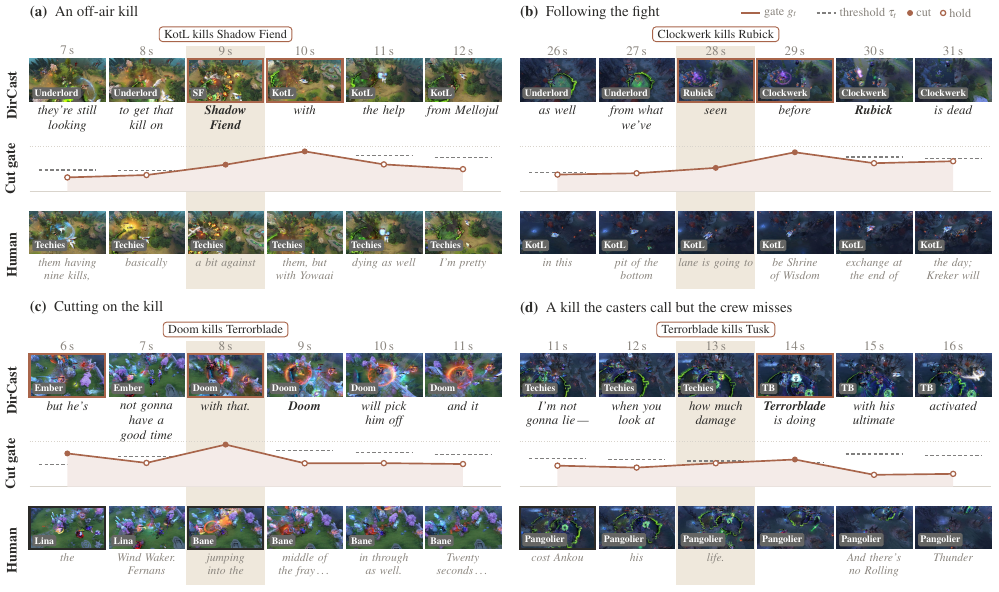}
    \caption{Four kills that DirCast shows and the human crew misses. Each panel shows DirCast's program and commentary (top), the cut gate against its threshold (middle), and the human broadcast (bottom). Framed shots are cuts, and the shaded column marks the kill: (a)~KotL kills Shadow Fiend off-air, (b)~Clockwerk kills Rubick, (c)~Doom kills Terrorblade, and (d)~Terrorblade kills Tusk.}
    \label{fig:qual}
\end{figure}

\needspace{5\baselineskip}
\paragraph{Qualitative analysis.}
Figure~\ref{fig:qual} shows four of the 19 off-air kills that DirCast puts on air and names, chosen among those where the human crew shows an uninvolved hero. In each case the cut gate stays below its threshold while the action is elsewhere, so DirCast holds its shot; in the second of the kill the gate rises above the threshold and DirCast cuts to the fight, often first to the victim and then to the killer as it moves. The commentary names a hero of the fight within about two seconds of the kill: Shadow Fiend as KotL takes him down off-air in (a), Rubick as Clockwerk finishes him while DirCast follows the fight in (b), Doom as he picks off Terrorblade on the kill in (c), and Terrorblade as his ultimate kills Tusk in (d). After each cut the threshold rises, so DirCast holds the new shot instead of flickering between feeds. The human crew misses all four kills, staying on an uninvolved hero such as Techies in (a), KotL in (b), Bane in (c) and Pangolier in (d); in (d) the casters call the kill although the broadcast never shows it, so the official feed shows none of the deaths as they happen.

%% file: sections/tables/direct.tex
\definecolor{oursbg}{HTML}{EEF2F7}
\definecolor{refgray}{HTML}{8A8A8A}
\newcolumntype{Y}{>{\centering\arraybackslash}X}

\begin{table}[t]
\centering
\caption{Test-split directing. A cut is correct if within two seconds of a human cut and to the same feed. Fight coverage measures seconds with a fighting hero on air. Cuts/min should approach the human rate of 7.9. DirCast averages three seeds. Best bold, second underlined.}
\label{tab:director}
\setlength{\tabcolsep}{4pt}
\small
\begin{tabularx}{\textwidth}{l YYY YYY YY}
\toprule
& \multicolumn{3}{c}{Feed agreement} & \multicolumn{3}{c}{Cuts} & \multicolumn{2}{c}{Coverage} \\
\cmidrule(lr){2-4} \cmidrule(lr){5-7} \cmidrule(lr){8-9}
Director & Top-1 $\uparrow$ & Tol. $\uparrow$ & Any-vis. $\uparrow$ & P $\uparrow$ & F1 $\uparrow$ & /min & Kill $\uparrow$ & Fight $\uparrow$ \\
\midrule
\multicolumn{9}{l}{\textit{Rules}} \\
Random (human rate)   & 10.6 & 12.9 & 20.1 & 6.4 & 5.8 & 6.9 & 20.9 & 19.1 \\
Static                & 17.1 & 20.4 & 23.0 & 0.0 & 0.0 & 0.0 & 20.9 & 19.2 \\
Action rule           & \underline{25.4} & \underline{30.3} & \underline{36.1} & \underline{13.9} & 11.5 & 5.8 & 78.5 & \textbf{62.4} \\
\midrule
\multicolumn{9}{l}{\textit{Zero-shot VLMs}} \\
Qwen3-VL-8B           & 17.6 & 21.4 & 28.6 & 8.9 & 12.7 & 20.6 & 74.6 & 32.6 \\
Qwen3-VL-32B          & 11.2 & 13.9 & 21.7 & 8.6 & 7.4 & 6.3 & 67.8 & 30.6 \\
InternVL3.5-8B        & 19.1 & 22.9 & 28.9 & 6.2 & 10.8 & 56.3 & 72.9 & 24.1 \\
InternVL3.5-38B       & 19.0 & 22.8 & 27.7 & 6.6 & 10.0 & 25.3 & 81.9 & 28.1 \\
LLaVA-OV-7B           & 9.2 & 12.0 & 18.6 & 6.5 & 9.2 & 20.5 & 41.2 & 18.8 \\
Mistral-Small-3.2-24B & 13.6 & 17.0 & 24.2 & 6.1 & 9.3 & 26.7 & 59.3 & 23.3 \\
Kimi-K2.6             & 16.7 & 20.4 & 28.4 & 8.2 & \underline{13.7} & 42.4 & 81.4 & 30.1 \\
GPT-5                 & 13.5 & 16.7 & 24.8 & 6.7 & 11.6 & 51.7 & \underline{85.3} & 33.0 \\
\midrule
\rowcolor{oursbg}
DirCast (ours)        & \textbf{27.0} & \textbf{32.0} & \textbf{41.8} & \textbf{15.8} & \textbf{14.1} & 6.7 & \textbf{89.1} & \underline{41.1} \\
\midrule
\textcolor{refgray}{Human director} &
\textcolor{refgray}{100} & \textcolor{refgray}{100} & \textcolor{refgray}{100} &
\textcolor{refgray}{100} & \textcolor{refgray}{100} & \textcolor{refgray}{7.9} &
\textcolor{refgray}{50.3} & \textcolor{refgray}{37.4} \\
\bottomrule
\end{tabularx}
\end{table}

%% file: sections/tables/commentary.tex
\begin{table}[htb]
\centering
\caption{Streaming commentary on the test split under the Proact-VL protocol, judged by GPT-5. CC is the win rate against GPT-4o. $^\dagger$Fine-tuned on our training split. DirCast is the mean of three seeds.}
\label{tab:commentary}
\setlength{\tabcolsep}{3pt}
\small
\begin{tabularx}{\textwidth}{l l YYc YYY YYY c}
\toprule
& & & & & \multicolumn{3}{c}{LiveU} & \multicolumn{3}{c}{FinalQ} & \\
\cmidrule(lr){6-8}\cmidrule(lr){9-11}
Method & Venue & CC $\uparrow$ & F1 $\uparrow$ & TimeDiff $\downarrow$ & Time & Rate & TextU & Fidelity & Contin. & Subst. & PAUC $\uparrow$ \\
\midrule
\multicolumn{12}{l}{\textit{Zero-shot, human program}} \\
LiveCC-7B-Base~\cite{chen2025live}           & CVPR'25    & \underline{19.38} & 64.61 & 8.32  & 5.11 & 4.99 & 4.53 & 4.56 & 3.17 & 2.36 & 0.008 \\
LiveCC-7B-Instruct~\cite{chen2025live}       & CVPR'25    & 17.73 & 79.95 & 1.84  & 5.75 & 6.15 & 5.69 & 4.01 & 3.19 & \underline{2.59} & \underline{0.015} \\
Dispider~\cite{qian2025dispider}             & CVPR'25    & 6.70  & 0.69  & 30.64 & 1.36 & 1.94 & 1.64 & 4.60 & 1.34 & 1.15 & 0.000 \\
MMDuet~\cite{wang2025videollm}               & EMNLP'25   & 3.81  & 16.99 & 9.79  & 3.13 & 1.87 & 2.99 & \textbf{5.74} & 2.25 & 1.52 & 0.001 \\
LiveStar-8B~\cite{yang2026livestar}          & NeurIPS'25 & 1.86  & 16.69 & 3.03  & 2.52 & 3.12 & \underline{5.91} & 3.89 & 2.44 & 1.96 & 0.001 \\
StreamingVLM~\cite{xu2026streamingvlm}       & ICLR'26    & 14.64 & 65.47 & 1.16  & \underline{5.97} & \underline{6.43} & 4.73 & 4.01 & 3.10 & 2.43 & 0.008 \\
MMDuet2~\cite{wang2026mmduet2}               & ICLR'26    & 3.71  & 26.18 & 5.03  & 3.87 & 3.40 & 5.19 & 4.10 & 2.33 & 1.89 & 0.001 \\
Proact-VL\textsubscript{LiveCC}~\cite{yan2026proact} & ICML'26 & 16.39 & \underline{83.20} & \underline{0.34} & 5.75 & 5.54 & 5.54 & 4.15 & 3.18 & 2.38 & 0.008 \\
Proact-VL\textsubscript{Qwen3}~\cite{yan2026proact}  & ICML'26 & 17.73 & \textbf{83.21} & \underline{0.34} & 5.79 & 5.49 & 5.63 & 4.54 & \underline{3.38} & 2.48 & 0.012 \\
\midrule
\multicolumn{12}{l}{\textit{Trained on our data, human program}} \\
LiveCC-7B-FT$^\dagger$~\cite{chen2025live}   & CVPR'25    & 14.85 & 13.41 & 16.22 & 2.54 & 3.32 & 3.26 & \underline{5.39} & 2.59 & 1.53 & 0.007 \\
Proact-VL\textsubscript{Qwen3}-FT$^\dagger$~\cite{yan2026proact} & ICML'26 & 18.25 & 83.08 & \underline{0.34} & 5.39 & 4.20 & 4.52 & 5.04 & 3.10 & 2.11 & 0.014 \\

\midrule
\multicolumn{12}{l}{\textit{Trained on our data, self-directed program}} \\
\rowcolor{oursbg}
DirCast (ours)                               &            & \textbf{29.21} & 78.22 & \textbf{0.30} & \textbf{6.57} & \textbf{6.50} & \textbf{5.92} & 4.97 & \textbf{4.00} & \textbf{2.95} & \textbf{0.024} \\
\bottomrule
\end{tabularx}
\end{table}

%% file: sections/tables/abl_arc.tex
\begin{table}[htb]
\definecolor{oursbg}{HTML}{EEF2F7}
\centering
\caption{Joint versus separate models. TD is TimeDiff.}
\label{tab:joint}
\setlength{\tabcolsep}{3pt}
\footnotesize
\begin{tabularx}{\linewidth}{ll*{6}{>{\centering\arraybackslash}X}}
\toprule
& & \multicolumn{2}{c}{Cost} & Dir. & \multicolumn{3}{c}{Commentary} \\
\cmidrule(lr){3-4}\cmidrule(lr){5-5}\cmidrule(lr){6-8}
Director & Comm. & GB$\downarrow$ & Time$\downarrow$ & Top-1$\uparrow$ & CC$\uparrow$ & F1$\uparrow$ & TD$\downarrow$ \\
\midrule
Dir-only & Com-only & 37.8 & 0.53 & 25.17 & 24.36 & 76.73 & 0.31 \\
Ours     & Com-only & 37.8 & 0.56 & \textbf{26.56} & 29.04 & 76.64 & 0.31 \\
\rowcolor{oursbg}
Ours     & Ours     & \textbf{18.9} & \textbf{0.32} & \textbf{27.00} & \textbf{29.21} & \textbf{78.54} & \textbf{0.30} \\
\midrule
\textcolor{gray}{Human} & Com-only & -- & -- & -- & 25.67 & 76.23 & 0.33 \\
\textcolor{gray}{Human} & Ours     & -- & -- & -- & 27.84 & 77.48 & \textbf{0.30} \\
\bottomrule
\end{tabularx}
\end{table}

%% file: sections/tables/componets.tex
\begin{table}[htb]
\definecolor{oursbg}{HTML}{EEF2F7}
\centering
\caption{Component ablations on the full test split.}
\label{tab:ablation}
\setlength{\tabcolsep}{4pt}
\small
\begin{tabularx}{\linewidth}{l*{4}{>{\centering\arraybackslash}X}}
\toprule
& \multicolumn{2}{c}{Directing}
& \multicolumn{2}{c}{Commentary} \\
\cmidrule(lr){2-3}\cmidrule(lr){4-5}
Variant
& Top-1 $\uparrow$
& Any-vis. $\uparrow$
& CC $\uparrow$
& TimeDiff $\downarrow$ \\
\midrule
\rowcolor{oursbg}
DirCast
& \textbf{27.00} & \underline{41.80}
& \textbf{29.21} & \textbf{0.30} \\
w/o hold-or-switch pointer
& 21.65 & 35.93
& -- & \underline{0.31} \\
w/o state tokens
& 22.59 & 37.98
& \textbf{29.21} & 0.32 \\
w/o program rollouts
& \underline{22.90} & 38.01
& \underline{28.35} & \textbf{0.30} \\
w/o event-anchored targets
& 26.10 & \textbf{43.06}
& -- & 0.32 \\
\bottomrule
\end{tabularx}
\end{table}

%% file: sections/tables/memory.tex
\begin{table}[htb]
\centering
\caption{Streaming eight 288-second windows on one GPU. Time: compute per video second; KV: final cache length.}
\label{tab:longstream}
\small
\begin{tabular*}{\columnwidth}{@{\extracolsep{\fill}} l ccc cccc @{}}
\toprule
& \multicolumn{3}{c}{Time (s)} & & & & \\
\cmidrule(lr){2-4}
Context & Mean $\downarrow$ & P95 $\downarrow$ & Last 36\,s $\downarrow$ & KV $\downarrow$ & GB $\downarrow$ & Top-1 $\uparrow$ & Speech \\
\midrule
Every view kept      & 0.60 & 1.05 & 0.86 & 92.6k & 32.0 & 23.7 & 67.2 \\
Program memory       & 0.42 & 0.68 & 0.55 & 47.6k & 25.1 & \underline{24.6} & 68.1 \\
\quad + 72\,s horizon & \underline{0.36} & \underline{0.58} & \underline{0.38} & \underline{12.8k} & \underline{19.8} & \underline{24.6} & 70.1 \\
\quad + 36\,s horizon & \textbf{0.29} & \textbf{0.52} & \textbf{0.29} & \textbf{6.9k} & \textbf{18.9} & \textbf{25.1} & 76.7 \\
\bottomrule
\end{tabular*}
\end{table}

%% file: sections/conclusion.tex
\section{Conclusion}
\label{sec:conclusion}

\begingroup
\emergencystretch=1em
\looseness=-1
We introduced live esports broadcast production, where a model receives every player's camera feed and must choose the on-air shot and commentate using only past observations. DirCast addresses both tasks with a single streaming video language model. A hold-or-switch director selects the on-air shot, commentary follows its choice, and program-only context bounds the cost of long streams while enabling real-time operation on one GPU. Trained on human broadcasts with perturbed programs and event-anchored targets, DirCast outperforms rule-based and zero-shot directors including GPT-5, shows more kills on air than human crews, and wins more comparisons than eleven streaming commentators watching human broadcasts. Together, these results establish the feasibility of unified, real-time esports production with a single model.
\par
\endgroup

\section*{Acknowledgments}
This study was carried out using the TSUBAME4.0 supercomputer at Institute of Science Tokyo.

%% file: sections/appendix_dataset.tex
\section{Dataset Construction}
\label{sec:app_data}

This section describes how the dataset is built: the director data (Section~\ref{sec:app_director}), the commentary (Section~\ref{sec:app_commentary}), and the cleaning applied to both (Section~\ref{sec:app_clean}).

\subsection{Director Data}
\label{sec:app_director}

\paragraph{Source data.}
We build a multi-camera director dataset from 27 professional \emph{Dota~2} matches (EPL Masters and BLAST Slam qualifiers, 2026). For each match we use two public sources: the official broadcast video (VOD), which is the only record of the human director's output and of the casters' voices, and the official replay file (\texttt{.dem}), which contains the complete game state at 30 ticks per second. Player and hero metadata are taken from OpenDota.

\paragraph{Broadcast-to-replay alignment.}
The broadcast and the replay share the in-game clock shown on the HUD. We OCR the HUD clock from the VOD and fit a linear mapping between VOD time and the game clock, with a residual of about 0.3\,s. Because broadcasts often cover only part of a match, and some archive records point to the wrong game, we additionally OCR the hero names on the HUD and keep only VOD stretches whose hero set matches the replay. Game pauses require special care: replay ticks and the broadcast keep running during a pause while the game clock stops, so the mapping is defined on replay ticks with the paused ticks subtracted. We validate the alignment with a consistency check: heroes whose name tags are visible together in a broadcast frame must be close to each other in the replay at the mapped tick. All 27 matches pass this check at zero time offset.

\paragraph{Candidate camera feeds.}
Candidate feeds are rendered with the Dota~2 client itself. Each replay is played back in a headless X server and recorded by ten spectator clients in parallel, one per player, each in the client's \emph{Hero Chase} mode following that player's hero with the HUD hidden. Clients are driven through the console interface; capture starts and ends at fixed replay ticks and every frame is tagged with its tick, so all ten feeds are synchronized with the replay and with the broadcast. Feeds are captured at $1280\times720$ and 30\,fps and stored at $640\times360$; training clips use one frame per second resized to $288\times288$. Each match is rendered in 300\,s windows covering the broadcast-aligned part of the game.

\paragraph{Director labels.}
The label for each second is the hero the broadcast is looking at, which maps to one of the ten feeds. We obtain it from the broadcast frame itself: the spectator HUD prints a name tag above every visible hero, which we OCR and match to the match roster, and the HUD also shows the currently selected unit. The label is the selected hero if it is on screen, otherwise the visible hero nearest the frame centre, with a 2\,s hysteresis on switches and up to 4\,s of carry-over when no tag is readable. Every label records which of these cases produced it, which the cleaning in Section~\ref{sec:app_clean} uses. Every second also receives a broadcast state (live, replay, production, paused, no coverage) derived from the clock OCR and a gameplay cue. We additionally store the set of all heroes on screen in each second, which supports an \emph{any-of-visible} evaluation, since several feeds can show the same scene.

\subsection{Commentary}
\label{sec:app_commentary}

\paragraph{Audio and transcription.}
For every rendered window we cut the broadcast audio at the same aligned VOD times as the director labels, so the commentary shares the time axis of the feeds and the labels, and resample it to 16\,kHz mono. The casters' speech is transcribed with Whisper (large-v3-turbo) in English with voice activity detection, beam size 5 and word-level timestamps. Each window is decoded without conditioning on the previous text, which avoids repetition loops over long audio, and with a vocabulary prompt that lists the heroes and players of the match, which biases the decoder towards the names the casters use. Every word is assigned to the second in which it starts, which gives the per-second commentary targets. Of the 2{,}096 training clips, 1{,}984 have broadcast audio, and about 8\% of the transcribed words have an ASR confidence below 0.5.

\paragraph{Polishing.}
Raw ASR misspells hero and player names and keeps fillers and repetitions. Such text would teach the model wrong names and would penalise correct names at evaluation. We therefore follow the data pipeline of Proact-VL~\cite{yan2026proact}, which merges ASR segments into sentences and polishes them with a language model. Consecutive Whisper segments are merged into a sentence until end punctuation, a pause of more than 1\,s, 30 words or 12\,s. Sentences are then rewritten in batches of twelve by GPT-5-mini with a match-specific prompt. The prompt lists both teams and their players and heroes, and asks the model to correct hero, player, item and ability names, to remove fillers and repetitions, and to keep the casters' own words, word order and meaning. Sentences that the casters leave unfinished stay unfinished, and the model may not add who did what, numbers or outcomes. Proact-VL uses DeepSeek-V3.2 for this step. We use GPT-5-mini because GPT-4o is the opponent and GPT-5 the judge of the commentary evaluation, so neither of them sees the references beforehand.

\paragraph{Guard and timing.}
A rewrite is accepted only if its word count lies between 70\% and 140\% of the ASR sentence and at least half of its words appear in the ASR sentence; otherwise we keep the ASR text with fillers removed. This guard prevents the language model from inventing content. Polishing keeps the timing of speech: the polished words of a sentence take the start times of the original words in order and are bucketed into seconds by the same rule as the raw transcript. Polishing changes the text of 38\% of the seconds with speech and removes 2\% of the words, mostly fillers and repetitions.

\paragraph{Use.}
The polished commentary is the training target of DirCast and of the two fine-tuned baselines, and it is the reference for every model at test time, as in the Proact-VL protocol.

\subsection{Cleaning}
\label{sec:app_clean}

Table~\ref{tab:app_clean} follows every clip-second through the cleaning steps.

\paragraph{Broadcast state.}
Only seconds in which the broadcast shows live gameplay are director decisions. Replays, production shots such as the desk or the drafting screen, pauses and seconds without broadcast coverage are removed, which keeps 87\% of the clip-seconds.

\paragraph{Label quality.}
Labels that come from the HUD selection alone, without the selected hero on screen, are dropped, since only 52\% of them agree with the hero positions in the replay. The remaining labels form the training mask.

\paragraph{Verification.}
Carried-over labels are the noisiest, because the broadcast may cut while no name tag is readable. We verify all 15{,}951 carried-over seconds with a vision--language model (GPT-6 Astra). It sees the broadcast frame and four candidate feeds, the labelled feed among them, and answers which of them shows the same scene; a label is confirmed if the answer is the labelled feed. 62\% are confirmed, and the rejected seconds are removed from the training mask, which gives the \emph{verified} mask used for training. On a manually audited sample the model's verdicts agree with the human audit in 43 of 45 cases.

\paragraph{Strict evaluation subset.}
Evaluation uses only labels that are observed on screen in that very second, that is, the selected hero with its tag visible or the visible hero nearest the frame centre, without hysteresis or carry-over. This \emph{strict} mask keeps 69\% of the live test seconds.

\paragraph{Commentary.}
Commentary is cleaned by the polishing and guard described above. Windows without broadcast audio keep their director labels but contribute no commentary targets.

\paragraph{Splits.}
Matches are split as a whole into 20 training, 3 validation and 4 test matches, so no game, broadcast or caster utterance is shared between training and test. The validation matches are used to calibrate the base thresholds of the decoding rules.

\subsection{Statistics}

The dataset contains 15.2 hours of aligned ten-feed footage, of which 13.1 hours are live broadcast and 10.3 hours carry verified labels (7.9 hours in the strict subset). It is packaged as 2{,}912 clips of 36\,s (stride 18\,s), split as in Table~\ref{tab:app_dataset}. Casters speak in 64\% of the training clip-seconds and 72\% of the test clip-seconds. The director switches feeds every few seconds, so a persistence baseline that repeats the previous label reaches about 80\% per-second accuracy; we therefore report switch-second accuracy separately.

\begin{table}[h]
\centering
\small
\setlength{\tabcolsep}{3.5pt}
\begin{tabular}{lrrrrrr}
\toprule
Split & Matches & Clips & Verified clip-s & Strict clip-s & Speech & Words \\
\midrule
Train & 20 & 2{,}096 & 50{,}283 & 35{,}815 & 64\% & 142.0k \\
Val   & 3  & 320     & 9{,}627  & 7{,}357  & 50\% & 16.3k \\
Test  & 4  & 496     & 14{,}630 & 11{,}283 & 72\% & 38.0k \\
\bottomrule
\end{tabular}
\caption{Dataset statistics. Clips are 36\,s long with a stride of 18\,s, so clip-seconds and words count overlapping clips; unique footage totals 15.2\,h. Speech is the share of clip-seconds with polished commentary.}
\label{tab:app_dataset}
\end{table}

\begin{table}[h]
\centering
\small
\setlength{\tabcolsep}{4pt}
\begin{tabular}{lrrr}
\toprule
Clip-seconds & Train & Val & Test \\
\midrule
All                         & 75{,}456 & 11{,}520 & 17{,}856 \\
Live broadcast              & 63{,}374 & 11{,}143 & 16{,}238 \\
\quad without HUD-only labels (training mask) & 59{,}574 & 10{,}977 & 15{,}894 \\
\quad after verification (verified mask)      & \textbf{50{,}283} & 9{,}627 & 14{,}630 \\
Observed on screen (strict mask)              & 35{,}815 & 7{,}357 & \textbf{11{,}283} \\
\bottomrule
\end{tabular}
\caption{Cleaning funnel. Bold marks the seconds used for training and for evaluation. Clip-seconds count overlapping clips.}
\label{tab:app_clean}
\end{table}

%% file: sections/appendix_qualitative.tex
\section{Additional qualitative examples}
\label{app:qualitative}
\begin{figure}[p]
    \centering
    \includegraphics[width=\textwidth,height=0.92\textheight,keepaspectratio]{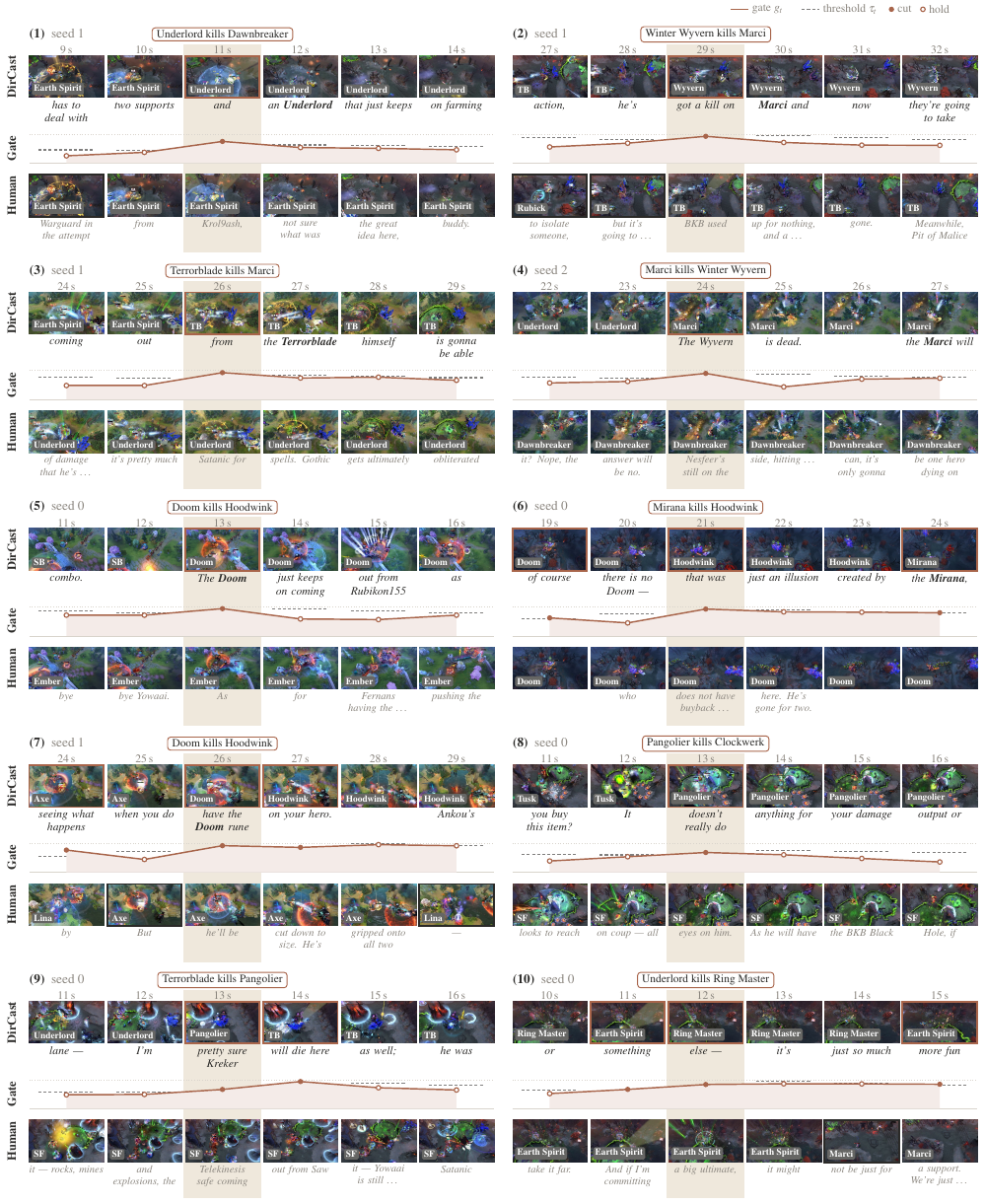}
    \caption{Additional qualitative examples of off-air kills that DirCast shows and names (1--10 of 20).}
    \label{fig:appendix_qual_p1}
\end{figure}

\begin{figure}[p]
    \centering
    \includegraphics[width=\textwidth,height=0.92\textheight,keepaspectratio]{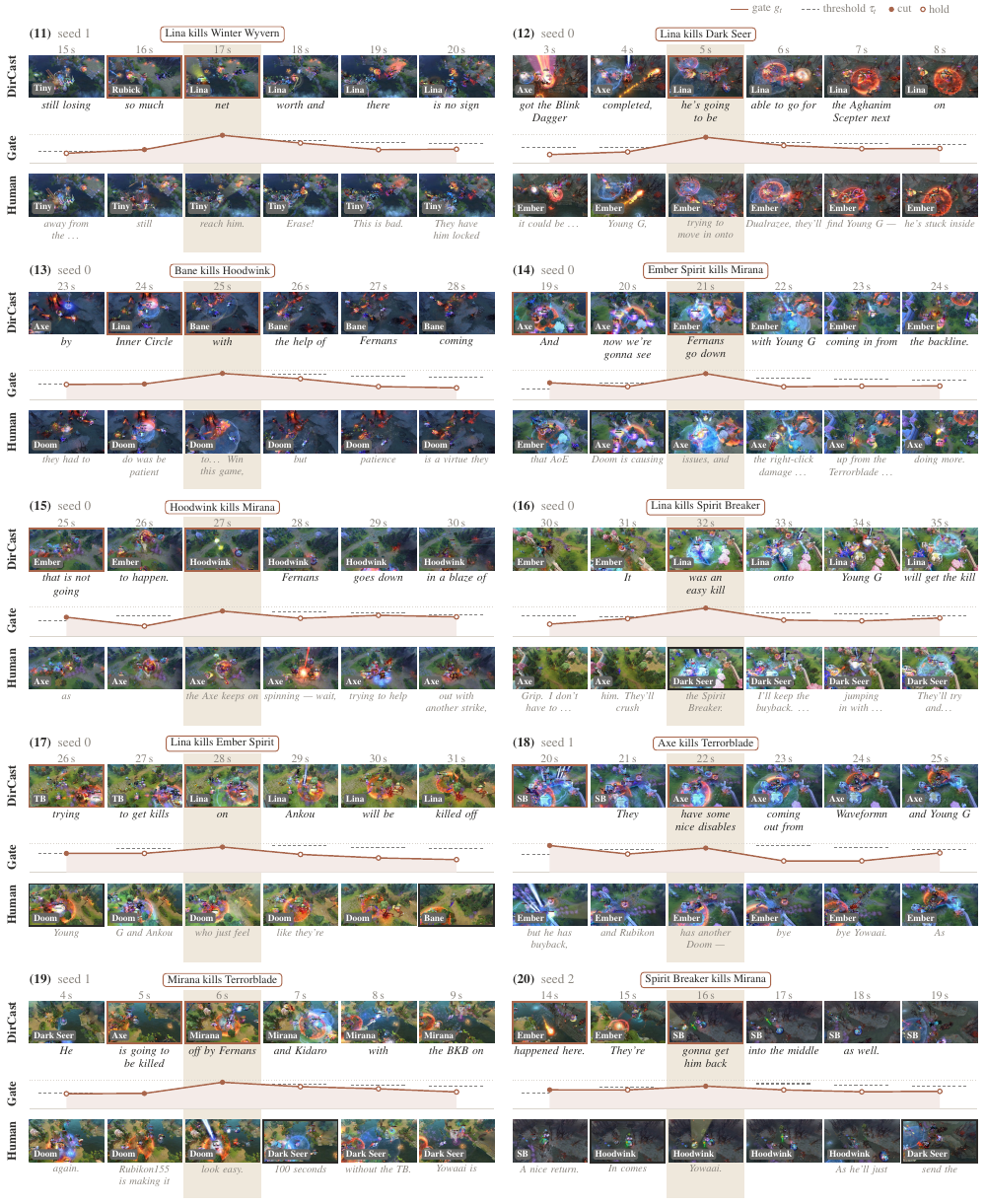}
    \caption{Additional qualitative examples of off-air kills that DirCast shows and names (11--20 of 20).}
    \label{fig:appendix_qual_p2}
\end{figure}

We show twenty more off-air kills that DirCast puts on air and names, ten in Figure~\ref{fig:appendix_qual_p1} and ten in Figure~\ref{fig:appendix_qual_p2}, drawn from the test matches and across the random seeds used in the main experiments. The examples cover a range of heroes and situations, from single-target pickoffs to crowded team fights, and each is chosen among the off-air kills where the human crew shows an uninvolved hero, so they test the director on moments it has to find on its own.

Each panel follows the format of Figure~\ref{fig:qual}. The label at the top names the kill and the seed. The top row is DirCast's program, one on-air thumbnail per second with the words it speaks printed underneath; the middle row plots the cut gate against its threshold, with markers for the seconds it cuts and the seconds it holds; and the bottom row is the human broadcast over the same seconds. The shaded column marks the second of the kill.

Across the examples DirCast shows the behaviour described in Section~\ref{sec:analysis}: it holds its shot while the gate stays below the threshold, cuts to the fight as the gate rises in the second around the kill, often moving from the victim to the killer as the fight develops, and names a hero of the fight within about two seconds. After the cut the threshold rises, so it holds the new shot instead of flickering between feeds. By construction the human crew is on an uninvolved hero in every panel, and it shows the kill late or not at all, so these decisive seconds never reach the official broadcast even though DirCast puts them on air and calls them.

